\documentclass[11pt]{arxform}
\usepackage{graphicx}
\usepackage{multirow}
\usepackage{todonotes}
\usepackage{stmaryrd}
\usepackage{amsmath}
\usepackage{amssymb}
\usepackage{hyperref}
\usepackage{subfig}

\usepackage{times}
\usepackage{multirow,color}
\usepackage{amsmath} 
\renewcommand{\vec}[1]{\boldsymbol{#1}}

\newcommand{\on}{\operatorname}

\usepackage{algorithm}
\usepackage{algorithmic}
\usepackage{stmaryrd}
\usepackage{amsfonts}
\usepackage{dsfont}
\usepackage{tabularx}

\begin{document}
	\title{Towards Analogy-Based Explanations\\ in Machine Learning\thanks{Draft of an article in Proc.\ MDAI 2020, 17th Int.\ Conf.\ on Modeling Decisions for Artificial Intelligence.}}
	%
	%
	\author{Eyke H{\"u}llermeier}
	%
	%
	
	\institute{Paderborn University\\
		Heinz Nixdorf Institute and Department of Computer Science\\
		Intelligent Systems and Machine Learning Group\\
		\email{eyke@upb.de}}
	\maketitle              
	\begin{abstract}
	Principles of analogical reasoning have recently been applied in the context of machine learning, for example to develop new methods for classification and preference learning. In this paper, we argue that, while analogical reasoning is certainly useful for constructing new learning algorithms with high predictive accuracy, is is arguably not less interesting from an interpretability and explainability point of view. More specifically, we take the view that an analogy-based approach is a viable alternative to existing approaches in the realm of explainable AI and interpretable machine learning, and that analogy-based explanations of the predictions produced by a machine learning algorithm can complement similarity-based explanations in a meaningful way. To corroborate these claims, we outline the basic idea of an analogy-based explanation and illustrate its potential usefulness by means of some examples.  
	\end{abstract}

	\section{Introduction}

Over the past couple of years, the idea of explainability and related notions such as transparency and interpretability have received increasing attention in artificial intelligence (AI) in general and machine learning (ML) in particular. This is mainly due to the ever growing number of real-world applications of AI technology and the increasing level of autonomy of algorithms taking decisions on behalf of people, and hence of the social responsibility of computer scientists developing these algorithms. Meanwhile, algorithmic decision making has a strong societal impact, which has led to the quest for understanding such decisions, or even to claiming a ``right to explanation'' \cite{good_eu17}. Explainability is closely connected to other properties characterizing a ``responsible'' or ``trusthworthy'' AI/ML, such as fairness, safety, robustness, responsibility, and accountability, among others.

Machine learning models, or, more specifically, the predictors induced by a machine learning algorithm on the basis of suitable training data, are not immediately understandable most of the time. This is especially true for the most ``fashionable'' class of ML algorithms these days, namely deep neural networks. On the contrary, a neural network is a typical example of what is called a ``black-box'' model in the literature: It takes inputs and produces associated outputs, often with high predictive accuracy, but the way in which the inputs and outputs are related to each other, and the latter are produced on the basis of the former, is very intransparent, as it involves possibly millions of mathematical operations and nonlinear transformations conducted by the network (in an attempt to simulate the neural activity of a human brain). A lack of transparency and interpretability is arguably less problematic for other ML methodology with a stronger ``white-box'' character, most notably symbol-oriented approaches such as rules and decision trees. Yet, even for such methods, interpretability is far from being guaranteed, especially because accurate models often require a certain size and complexity. For example, even if a decision tree might be interpretable in principle, a tree with hundreds of nodes will hardly be understandable by anyone.

The lack of transparency of contemporary ML methodology has triggered research that is aimed at improving the interpretability of ML algorithms, models, and predictions. In this regard, various approaches have been put forward, ranging from ``interpretability by design'', i.e., learning models with in-built interpretability, to model-agnostic explanations\,---\,a brief overview will be given in the next section. In this paper, we propose to add principles of \emph{analogical reasoning} \cite{gent_tm89} as another alternative to this repertoire. Such an approach is especially motivated by so-called example-based explanations, which refer to the notion of \emph{similarity}. Molnar \cite{moln_im} describes the blueprint of such explanations as follows: ``Thing B is similar to thing A and A caused Y, so I predict that B will cause Y
as well.'' In a machine learning context, the ``things'' are data entities (instances), and the causes are predictions. In (binary) classification, for example, the above pattern might be used to explain the learner's prediction for a query instance: A belongs to the positive class, and B is similar to A, hence B is likely to be positive, too. Obviously, this type of explanation is intimately connected to the nearest neighbor estimation principle \cite{cove_nn67}. 

Now, while similarity establishes a relationship between pairs of objects (i.e.\ tuples), an analogy involves four such objects (i.e.\ quadruples). The basic regularity assumption underlying analogical reasoning is as follows: Given objects $A$, $B$, $C$, $D$, if $A$ relates to $B$ as $C$ relates to $D$, then this ``relatedness'' also applies to the properties caused by these objects (for example, the predictions produced by an ML model). Several authors have recently elaborated on the idea of using analogical reasoning for the purpose of (supervised) machine learning \cite{BOUNHAS201736,ahmadi_huellermeier_aaai18,app}, though without raising the issue of interpretability. Here, we will argue that analogy-based explanations can complement similarity-based explanations in a meaningful way.

The remainder of the paper is organized as follows. In the next section, we give a brief overview of different approaches to interpretable machine learning. In Section 3, we provide some background on analogy-based learning\,---\,to this end, we recall the basics of a concrete method that was recently introduced in \cite{ahmadi_huellermeier_aaai18}. In Section 4, we elaborate on the idea on analogy-based explanations in machine learning, specifically focusing on classification and preference learning.	

\section{Interpretable Machine Learning}

In the realm of interpretable machine learning, two broad approaches are often distinguished. The first is to learn models that are inherently interpretable, i.e., models with in-built transparency that are interpretable by design. Several classical ML methods are put into this category, most notably symbol-oriented approaches like decision trees, but also methods that induce ``simple'' (typically linear) mathematical models, such as logistic regression. The second approach is to extract interpretable information from presumably intransparent ``black-box'' models.  Within this category, two subcategories can be further distinguished. 

In the first subcategory, \emph{global} approximations of the entire black-box model are produced by training more transparent ``white-box'' models as a surrogate. This can be done, for example, by using the black-box model as a teacher, i.e., to produce training data for the white-box model \cite{andr_sa95}.   
In the second subcategory, which is specifically relevant for this paper, the idea is to extract interpretable \emph{local} information, which only pertains to a restricted region in the instance space, or perhaps only to a single instance. In other words, the idea is to approximate a black-box model only locally instead of globally, which, of course, can be accomplished more easily, especially by simple models. Prominent examples of this approach are LIME \cite{ribe_ws16} and SHAP \cite{lund_au17}. These approaches are qualified as \emph{model agnostic}, because they use the underlying model only as a black-box that is queried for the purpose of data generation. 

In addition to generic, universally applicable methods of this kind, there are various methods for extracting useful information that are specifically tailored to certain model classes, most notably deep neural networks \cite{same_ea}. Such methods seek to provide some basic understanding of how such a network connects inputs with outputs. To this end, various techniques for making a network more transparent have been proposed, many of them based on the visualization of neural activities. 

Interestingly, the focus in interpretable machine learning has been very much on classification so far, while other ML problems have been considered much less. In particular, there is very little work on interpretable preference learning and ranking \cite{mpub218}. As will be argued later on, the idea of analogy-based explanation appears to be especially appealing from this point of view.

	\section{Analogy-Based Learning}\label{sec:able2rank}
	
In this section, we briefly recall the basic ideas of an analogy-based learning algorithm that was recently introduced in \cite{ahmadi_huellermeier_aaai18}. This will set the stage for our discussion of analogy-based explanation in the next section, and provide a basis for understanding the main arguments put forward there.

The mentioned approach proceeds from the standard setting of supervised learning, in which data objects (instances) are described in terms of feature vectors $\boldsymbol{x}  = (x_1, \ldots , x_d) \in \mathcal{X} \subseteq \mathbb{R}^d$. The authors are mainly interested in the problem of \emph{ranking}, i.e., in learning a ranking function $\rho$ that accepts any (query) subset $Q = \{ \boldsymbol{x}_1, \ldots , \boldsymbol{x}_n \} \subseteq \mathcal{X}$ of instances as input. As output, the function produces a ranking in the form of a total order of the instances, which can be represented by a permutation $\pi$, with $\pi(i)$ the rank of instance $\boldsymbol{x}_i$. Yet, the algorithmic principles underlying this approach can also be used for the purpose of classification. In the following, to ease explanation, we shall nevertheless stick to the case of ranking. 
	
\subsection{Analogical Proportions}

The approach essentially builds on the following inference pattern: If object $\boldsymbol{a}$ relates to object $\boldsymbol{b}$ as $\boldsymbol{c}$ relates to $\boldsymbol{d}$, and knowing that $\boldsymbol{a}$ is preferred to $\boldsymbol{b}$, we (hypothetically) infer that $\boldsymbol{c}$ is preferred to $\boldsymbol{d}$. This principle is formalized using the concept of analogical proportion \cite{Miclet2009}. For every quadruple of objects $\boldsymbol{a},\boldsymbol{b},\boldsymbol{c},\boldsymbol{d}$, the latter provides a numerical degree to which these objects are in analogical relation to each other. To this end, such a degree is first determined for each attribute value (feature) separately, and these degrees are then combined into an overall degree of analogy.   
	
	More specifically, for four values $a, b, c, d$ from an attribute domain $\mathbb{X}$, the quadruple $(a,b,c,d)$ is said to be in analogical proportion, denoted by $a:b::c:d$, if ``$a$ relates to $b$ as $c$ relates to $d$'', or formally:
	\begin{equation}\label{eq:ap}
	E \big( \mathcal{R}(a,b) , \mathcal{R}(c,d) \big) \, ,
	\end{equation}
	where the relation $E$ denotes the ``as'' part of the informal description. $\mathcal{R}$ can be instantiated in different ways, depending on the underlying domain $\mathbb{X}$: 
\begin{itemize}
\item	In the case of Boolean variables, where $\mathbb{X} = \{0,1\}$, there are $2^4=16$ instantiations of the pattern $a:b::c:d$, of which only the following 6 satisfy a set of axioms required to hold for analogical proportions: 
		\begin{center}
			\begin{tabular}{cccc}
				\hline
				\;$a$\; & \;$b$\; & \;$c$\; & \;$d$\; \\
				\hline
				0 & 0 & 0 & 0 \\
				0 & 0 & 1 & 1 \\
				0 & 1 & 0 & 1 \\
				1 & 0 & 1 & 0 \\
				1 & 1 & 0 & 0 \\
				1 & 1 & 1 & 1 \\
				\hline
			\end{tabular}
		\end{center}
	This formalization captures the idea that $a$ differs from $b$ (in the sense of being ``equally true'', ``more true'', or ``less true'', if the values 0 and 1 are interpreted as truth degrees) exactly as $c$ differs from $d$, and vice versa.
\item
	In the numerical case, assuming all attributes to be normalized to the unit interval $[0,1]$, the concept of analogical proportion can be extended on the basis of generalized logical operators \cite{BOUNHAS201736,dubois16}. In this case, the analogical proportion will become a matter of degree, i.e., a quadruple $(a,b,c,d)$ can be in analogical proportion \emph{to some degree} between 0 and 1. An example of such a proportion, with $\mathcal{R}$ being the arithmetic difference, i.e., $\mathcal{R}(a,b)=a-b$, is the following: 
	\begin{equation}\label{eq:v_a}
	v(a,b,c,d) =
		\begin{cases}
		1- | (a-b) - (c-d)| ,& \text{if } \on{sign}(a-b) = \on{sign}(c-d)\\
		0,              & \text{otherwise.}
		\end{cases}
		\end{equation}
	Note that this formalization indeed generalizes the Boolean case (where $a,b,c,d \in \{0,1 \}$). Another example is geometric proportions $\mathcal{R}(a,b)=a/b$.
	\end{itemize}
	To extend analogical proportions from individual values to complete feature vectors, the individual degrees of proportion can be combined using any suitable aggregation function, for example the arithmetic mean:
	$$
	v(\boldsymbol{a}, \boldsymbol{b} , \boldsymbol{c} , \boldsymbol{d}) = \frac{1}{d} \sum_{i=1}^d 
	v(a_i , b_i , c_i , d_i)  \, .
	$$

\subsection{Analogical Prediction}

\begin{figure} 
		\centering
		\includegraphics[width=0.6\columnwidth]{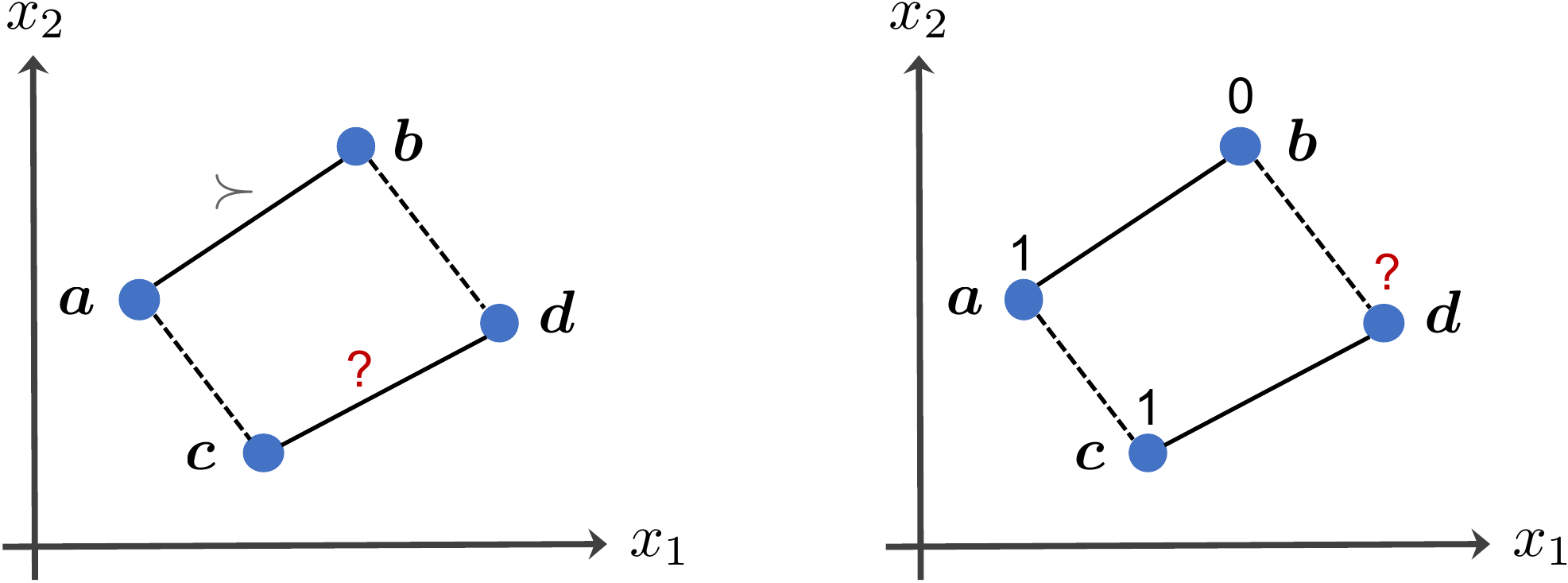}
		\caption{Illustration of analogy-based prediction: The four objects $\vec{a}$, $\vec{b}$, $\vec{c}$, $\vec{d}$ are in analogical proportion to each other. In the case of preference learning, the known preference $\vec{a} \succ \vec{b}$ would hence be taken as an indication that $\vec{c} \succ \vec{d}$ (left). Likewise, in the case of binary classification, knowing that $\vec{a}$ and $\vec{c}$ are positive while $\vec{b}$ is negative, analogical inference suggests that $\vec{d}$ is negative, too (right).}
		\label{fig:analogy}
	\end{figure}

The basic idea of analogy-based learning is to leverage analogical proportions for the purpose of \emph{analogical transfer}, that is, to transfer information about the target of prediction. In the case of preference learning, the target could be the preference relation between two objects $\boldsymbol{c}$ and $\boldsymbol{d}$, i.e., whether $\boldsymbol{c} \succ \boldsymbol{d}$ or $\boldsymbol{d} \succ \boldsymbol{c}$. Likewise, in the case of classification, the target could be the class label of a query object $\boldsymbol{d}$ (cf.\ Fig.\ \ref{fig:analogy} for an illustration).

In the context of preference learning, the authors in \cite{ahmadi_huellermeier_aaai18} realize analogical transfer in the style of the $k$-nearest neighbor approach: Given a query pair $(\boldsymbol{c} , \boldsymbol{d})$, they search for the tuples $(\boldsymbol{a}_i , \boldsymbol{b}_i)$ in the training data producing the $k$ highest analogies $\{\boldsymbol{a}_i : \boldsymbol{b}_i:: \boldsymbol{c} : \boldsymbol{d}\}_{i=1}^k$. Since the preferences between $\boldsymbol{a}_i$ and $\boldsymbol{b}_i$ are given as part of the training data, each of these analogies suggests either $\boldsymbol{c} \succ \boldsymbol{d}$ or $\boldsymbol{d} \succ \boldsymbol{c}$ by virtue of analogical transfer, i.e., each of them provides a vote in favor of the first or the second case. Eventually, the preference with the higher number of votes is adopted, or the distribution of votes is turned into an estimate of the probability of the two cases.

Obviously, a very similar principle could be invoked in the case of (binary) classification. Here, given a query instance $\boldsymbol{d}$, one would search for triplets $(\boldsymbol{a}_i , \boldsymbol{b}_i, \boldsymbol{c}_i)$ in the training data forming strong analogies $\boldsymbol{a}_i : \boldsymbol{b}_i :: \boldsymbol{c}_i : \boldsymbol{d}$, and again invoke the principle of analogical transfer to conjecture about the class label of $\boldsymbol{d}$. Each analogy will suggest the positive or the negative class, and the corresponding votes could then be aggregated in one way or the other.

\subsection{Feature Selection}\label{sec:fs}

Obviously, the feature representation of objects will have a strong effect on whether, or to what extent, the analogy assumption applies to a specific problem, and hence influence the success of the analogical inference principle. Therefore, prior to applying analogical reasoning methods, it could make sense to find an embedding of objects in a suitable space, so that the assumption of the above inference pattern holds true in that space. This is comparable, for example, to embedding objects in $\mathbb{R}^d$ in such a way that the nearest neighbor rule with Euclidean distance yields good predictions in a classification task.

In \cite{mpub394}, the authors address the problem of \emph{feature selection} \cite{guyo_ai03} in analogical inference, which can be seen as a specific type of embedding, namely a projection of the data from the original feature space to a subspace. By ignoring irrelevant or noisy features and restricting to the most relevant dimensions, feature selection can often improve the performance of learning methods. Moreover, feature selection is also important from an explainability point of view, because the representation of data objects in terms of meaningful features is a basic prerequisite for the interpretability of a machine learning model operating on this representation. In this regard, feature selection is also more appropriate than general feature embedding techniques. The latter typically produce new features in the form of (nonlinear) combinations of the original features, which lose semantic meaning and are therefore difficult to interpret. 
	
\section{Analogy-based Explanation}

To motivate an analogy-based explanation of predictions produced by an ML algorithm, let us again consider the idea of similarity-based explanation as a starting point. As we shall argue, the former can complement the latter in a meaningful way, especially because it refers to a different type of ``knowledge transfer''. As before, we distinguish between two exemplary prediction tasks, namely classification and ranking. This distinction is arguably important, mainly for the following reason: In classification, a property (class membership) is assigned to a \emph{single} object $\vec{x}$, whereas in ranking, a property (preference) is ascribed to a \emph{pair} of objects $(\vec{c}, \vec{d})$. Moreover, in the case of ranking, the property is in fact a relation, namely a binary preference relation. Thus, since analogy-based inference essentially deals with ``relatedness'', ranking and preference learning lends itself to analogy-based explanation quite naturally, perhaps even more so than classification.

For the purpose of illustration, we make use of a data set that classifies 172 scientific journals in the field of pure mathematics into quality categories $A^*$, $A$, $B$, $C$  \cite{beli_cb10}. Each journal is moreover scored in terms of 5 criteria, namely 
\begin{itemize}
\item cites: the total number of citations per year; 
\item IF: the well-known impact factor (average number of citations per article within
two years after publication);
\item II: the immediacy index measures how topical the articles published in a journal are (cites to articles in current calendar year divided by the number of articles published in that year);
\item articles: the total number of articles published;
\item half-line: cited half-life (median age of articles cited).
\end{itemize}
In a machine learning context, a classification task may consist of predicting the category of a journal, using the scores on the criteria as features. 
Likewise, a ranking task may consist of predicting preferences between journals, or predicting an entire ranking of several journals.


\subsection{Explaining Class Predictions}

\begin{figure} 
		\centering
		\includegraphics[width=0.7\columnwidth]{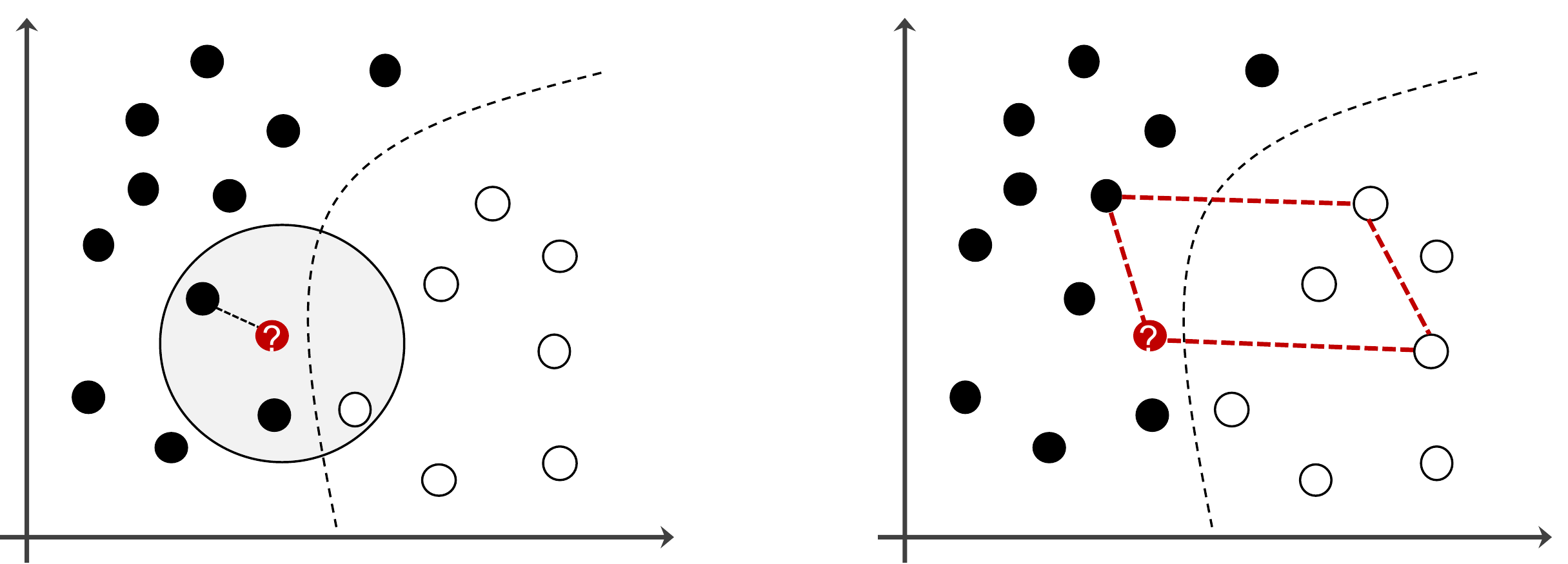}
		\caption{Left: Illustration of similarity-based explanation in a binary classification setting (black points are positive examples, white ones negative). The shaded circle indicated the neighborhood (of size 3) of the query instance (red point). The dashed line is a discriminant function that could have been induced by another (global) learning method. Right: Illustration of analogy-based explanation. Again, the query point is shown in red. Together with the three other points, it forms an analogical relationship.}
		\label{fig:nn}
	\end{figure}

Similarity-based explanations typically ``justify'' a prediction by referring to local (nearest neighbor) information in the vicinity of a query instance $\vec{x}$. In the simplest case, the nearest neighbor of $\vec{x}$ is retrieved from the training data, and its class label is provided as a justification (cf.\ Fig.\ \ref{fig:nn}): ``There is a case  $\vec{x}'$ that is similar to $\vec{x}$, and which belongs to class $y$, so $\vec{x}$ is likely to belong to $y$ as well.'' 
For example, there is another journal with similar scores on the different criteria, and which is ranked in category $A$, which explains why this journal is also put into this category.  

A slightly more general approach is to retrieve, not only the single nearest neighbor but the $k$ nearest neighbors, and to provide information about the distribution of classes among this set of examples. Information of that kind is obviously useful, as it conveys an idea of the confidence and reliability of a prediction. If many neighbors are all from the same class, this will of course increase the trust in a prediction. If, on the other side, the distribution of classes in the neighborhood is mixed, the prediction might be considered as uncertain, and hence the explanation as less convincing. 

Similarity- or example-based explanations of this kind are very natural and suggest themselves if a nearest neighbor approach is used by the learner to make predictions. It should be mentioned, however, that similarity-based explanations can also be given if predictions are produced by another type of model (like the discriminative model indicated by the dashed decision boundary in Fig.\ \ref{fig:nn}). In this case, the nearest neighbor approach serves as a kind of surrogate model. This could be justified by the fact that most machine learning methods do indeed obey the regularity assumption underlying similarity-based inference. For example, a discriminant function is more likely to assign two similar objects to the same class than to separate them, although such cases do necessarily exist as well. 

Obviously, a key prerequisite of the meaningfulness of similarity-based explanations is a meaningful notion of similarity, formalized in terms of an underlying similarity or distance function. This assumption is far from trivial, and typically not satisfied by ``default'' metrics like the Euclidean distance. Instead, a meaningful measure of distance needs to properly modulate the influence of individual features, because not all features might be of equal importance. Besides, such a measure should be able to capture interactions between features, which might not be considered independently of each other. 
For example, depending on the value of one feature, another feature might be considered more or less important (or perhaps completely ignored, as it does not apply any more). 
In other words, a meaningful measure of similarity may require a complex aggregation of the similarities of individual features \cite{mpub173}. Needless to say, this may hamper the usefulness of similarity-based explanations: If a user cannot understand why two cases are deemed similar, she will hardly accept a similar case as an explanation.

Another issue of the similarity-based approach, which brings us to the analogy-based alternative, is related to the difficulty of interpreting a \emph{degree} of similarity or distance. Often, these degrees are not normalized (especially in the case of distance), and therefore difficult to judge: How similar is similar? What minimal degree of similarity should be expected in an explanation? 
For example, when explaining the categorization of a journal as $B$ by pointing to a journal with similar properties, which is also rated as $B$, one may wonder whether the difference between them is really so small, or not perhaps big enough to justify a categorization as $A$.

An analogy-based approach might be especially interesting in this regard, as it explicitly refers to this distance, or, more generally, the relatedness between data entities. In the above example, an analogy-based explanation given to the manager of a journal rated as $B$ instead of $A$ might be of the following kind: ``There are three other journals, two rated $A$ ($\vec{a}$ and $\vec{c}$) and one rated $B$ ($\vec{b}$). The relationship between $\vec{a}$ and $\vec{b}$ is very much the same as the relationship between $\vec{c}$ and your journal, and $\vec{b}$ was also rated $B$.'' For example, $\vec{a}$ and $\vec{b}$ may have the same characteristics, except that $\vec{b}$ has 100 more articles published per year. The manager might now be more content and accept the decision more easily, because she understands that 100 articles more or less can make the difference. 



\begin{figure} 
		\centering
		\includegraphics[width=0.3\columnwidth]{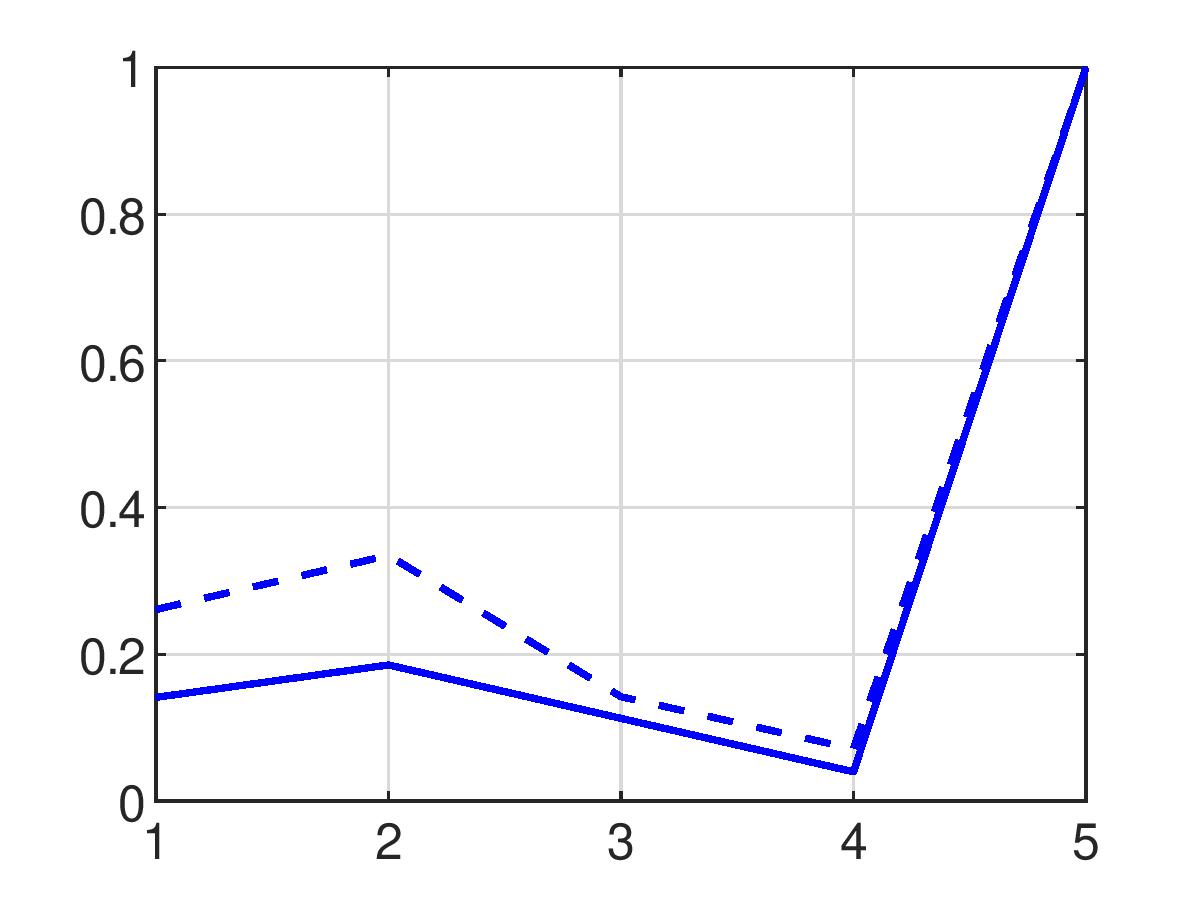}
		\includegraphics[width=0.3\columnwidth]{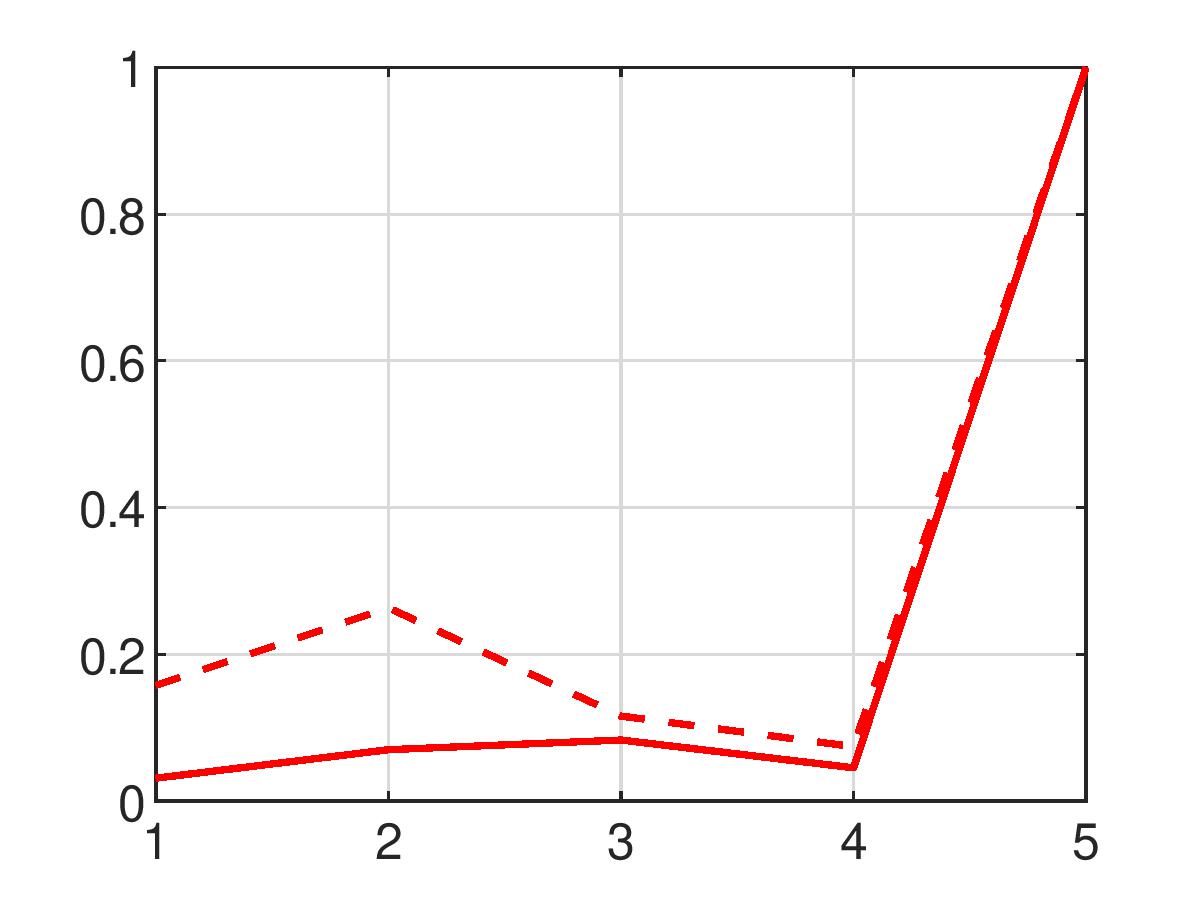}
		\caption{Example of an analogy $\vec{a}:\vec{b}::\vec{c}:\vec{d}$ in the journal data set, with $\vec{a}$ (dashed line) and $\vec{b}$ (solid) on the left panel, $\vec{c}$ (dashed line) and $\vec{d}$ (solid) on the right.}
		\label{fig:profiles}
	\end{figure}

A concrete example for the journal data is shown in Fig.\ \ref{fig:profiles}. Here, an analogy is found for a journal with (normalized) scores of 0.03, 0.06, 0.08, 0.04, 1 on the five criteria.
To explain why this journal is only put in category $C$ but not in $B$, three other journals $\vec{a}, \vec{b}, \vec{c}$ are found, $\vec{a}$ from category $A$, and $\vec{b}$ and $\vec{c}$ from category $B$, so that $\vec{a}:\vec{b}::\vec{c}:\vec{d}$ (the degree of analogical proportion is $\approx 0.98$). Note that the score profiles of $\vec{a}$ and $\vec{c}$ resp.\ $\vec{b}$ and $\vec{d}$ are not very similar themselves, which is not surprising, because $\vec{a}$ and $\vec{c}$ are from different categories. Still, the four journals form an analogy, in the sense that an $A$-journal relates to a $B$-journal as a $B$-journal relates to a $C$-journal.

Note that this type of explanation is somewhat related to the idea of explaining with \emph{counterfactuals} \cite{vanl_ic19}, although the cases forming an analogy are of course factual and not counterfactual. Nevertheless, an analogy-based explanation may give an idea of what changes of features might be required to achieve a change in the classification. On the other side, an analogy can of course also explain why a certain difference is \emph{not} enough. 
For example, explaining the rating as $B$ to a journal $\vec{d}$ by pointing to another journal $\vec{c}$ might not convince the journal manager, because she feels that, despite the similarity, her journal is still a bit better in most of the criteria. Finding an analogy $\vec{a}:\vec{b}::\vec{c}:\vec{d}$, with journals $\vec{a}$ and $\vec{b}$ also categorized as $B$, may then be helpful and convince her that the difference is not significant enough. 

\begin{figure} 
		\centering
		\includegraphics[width=0.4\columnwidth]{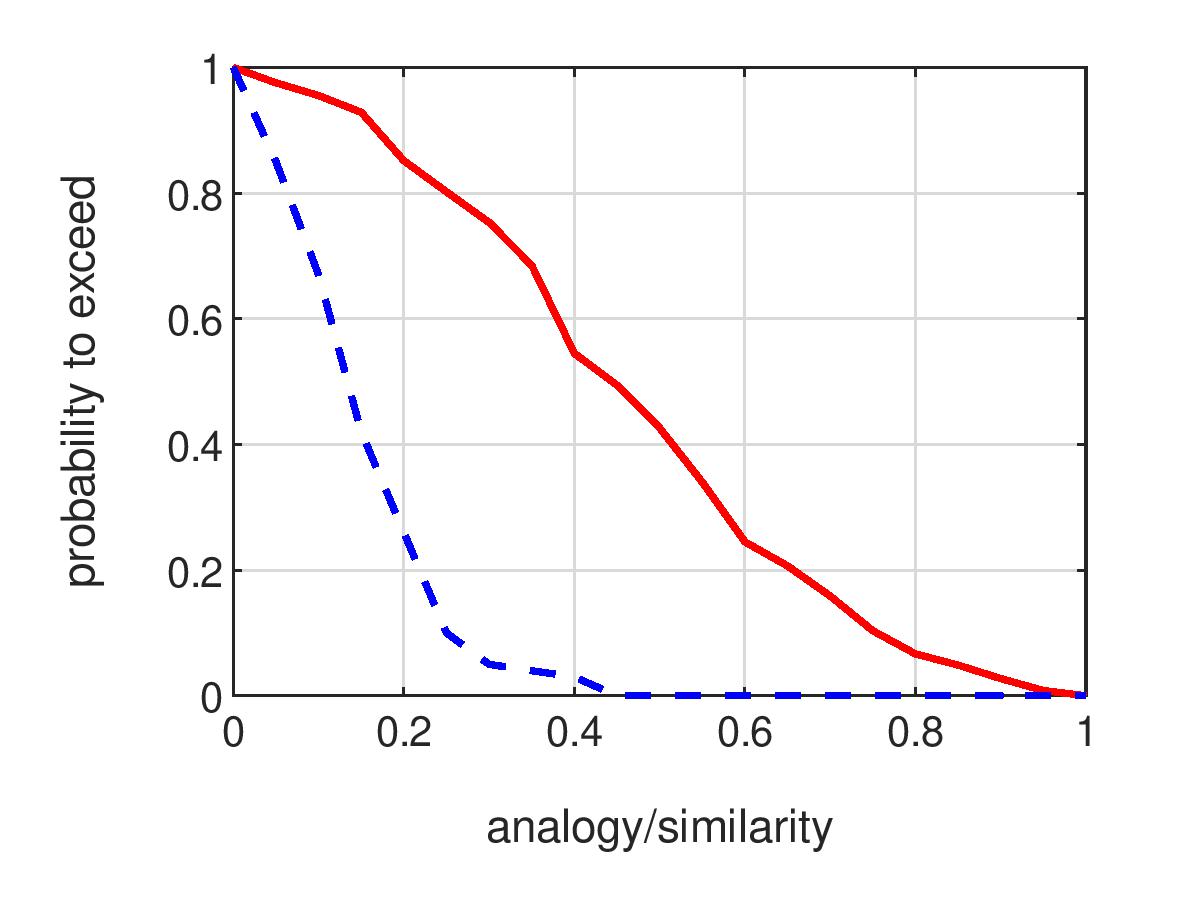}
		\caption{Decumulative analogy/similarity distribution: The solid line shows the average percentage of triplets (y-axis) $\vec{a}, \vec{b}, \vec{c}$ in the journal data set exceeding a degree of analogical proportion (x-axis) with a specific query $\vec{d}$. Likewise, the dashed line shows the average percentage of examples in the training data exceeding a degree of similarity ($1-L_1$-distance) with a query instance. As can be seen, the latter drops much faster than the former.}
		\label{fig:count}
	\end{figure}

Of course, just like the nearest neighbor approach, analogy-based explanations are not restricted to a single analogy. Instead, several analogies, perhaps sorted by their strength (degree of analogical proportion), could be extracted from the training data. In this regard, another potential advantage of an analogy-based compared to a similarity-based approach should be mentioned: While both approaches are local in the sense of giving an explanation for a specific query instance, i.e., local with respect to the \emph{explanandum}, the similarity-based approach is also local with respect to the \emph{explanans}, as it can only refer to cases in the vicinity of the query\,---\,which may or may not exist. The analogy-based approach, on the other side, is not restricted in this sense, as the explanans is not necessarily local. Instead, a triplet $\vec{a}, \vec{b}, \vec{c}$ forming an analogy with a query $\vec{d}$ can be distributed in any way, and hence offers many more possibilities for constructing an explanation (see also Fig.\ \ref{fig:count}). Besides, one should note that there are much more triplets than potential nearest neighbors (the former scales cubicly with the size of the training data, the latter only linearly).

Last but not least, let us mention that analogy-based explanations, just like similarity-based explanations, are in principle not limited to analogical learning, but could also be used in a model-agnostic way, and as a surrogate for other models.

\subsection{Explaining Preference Predictions}

In the case of classification, training data is given in the form of examples of the form $(\vec{x},y)$, where $\vec{x}$ is an element from the instance space $\mathcal{X}$ and $y$ a class label. In the case of preference learning, we assume training data in the form of pairwise preferences $\vec{a} \succ \vec{b}$, and the task is to infer the preferential relationship for a new object pair $(\vec{c}, \vec{d}) \in \mathcal{X} \times \mathcal{X}$ given as a query (or a ranking of more than two objects, in which case the prediction of pairwise preferences could be an intermediate step). How could one explain a prediction $\vec{c} \succ \vec{d}$? 

The similarity-based approach amounts to finding a ``similar'' preference $\vec{a} \succ \vec{b}$ in the training data. It is not immediately clear, however, what similarity of preferences is actually supposed to mean, even if a similarity measure on $\mathcal{X}$ is given. A natural solution would be a conjunctive combination: a preference $\vec{a} \succ \vec{b}$ is similar to $\vec{c} \succ \vec{d}$ if both $\vec{a}$ is similar to $\vec{c}$ and $\vec{b}$ is similar to $\vec{d}$. This requirement might be quite strong, so that finding similar preferences gets difficult. 

The analogy-based approach can be seen as a relaxation. Instead of requiring the objects to more or less coincide, they are only supposed to stand in a similar relationship to each other, i.e., $\mathcal{R}(\vec{a}, \vec{b}) \sim \mathcal{R}(\vec{c}, \vec{d})$. The relationship $\mathcal{R}$ \emph{can} mean similarity, but does not necessarily need to do so (as shown by the definition of $\mathcal{R}$ in terms of arithmetic or geometric proportions).

The explanation of preferences in terms of analogy appears to be quite natural. For the purpose of illustration, consider again our example: Why did the learner predict a preference $\vec{c} \succ \vec{d}$, i.e., that journal $\vec{c}$ is ranked higher (evaluated better than) journal $\vec{d}$? To give an explanation, one could find a preference $\vec{a} \succ \vec{b}$ between journals in the training data, so that $(\vec{a} , \vec{b})$ is in analogical proportion to $(\vec{c} , \vec{d})$. In the case of arithmetic proportions, this means that the feature values (ratings of criteria) of $\vec{a}$ deviate from the feature values of $\vec{d}$ in much the same way as those of $\vec{c}$ deviate from those of $\vec{d}$, and this deviation will then serve as an explanation of the preference. 


	\section{Conclusion and Future Work}

In this paper, we presented some preliminary ideas on leveraging the principle of analogy for the purpose of explanation in machine learning. This is essentially motivated by the recent interest in analogy-based approaches to ML problems, such as classification and preference learning, though hitherto without explicitly addressing the notion of interpretability. In particular, we tried to highlight the potential of an analogy-based approach to complement similarity-based (example-based) explanation in a reasonable way. 

Needless to say, our discussion is just a first step, and the evidence we presented in favor of analogy-based explanations is more of an anecdotal nature. In future work, the basic ideas put forward need to be worked out in detail, and indeed, there are many open questions to be addressed. For example, which analogy in a data set is best suited for explaining a specific prediction? In the case of analogy, the answer appears to be less straightforward than in the case of similarity, where more similarity is simply better than less. Also, going beyond the retrieval of a single analogy for explanation, how should one assemble a good composition of analogies? Last but not least, it would of course also be important to evaluate the usefulness of analogy-based explanation in a more systematic way, ideally in a user study involving human domain experts.

	%

\begin{thebibliography}{10}
\providecommand{\url}[1]{\texttt{#1}}
\providecommand{\urlprefix}{URL }
\providecommand{\doi}[1]{https://doi.org/#1}

\bibitem{mpub394}
{Ahmadi Fahandar}, M., H\"ullermeier, E.: Feature selection for analogy-based
  learning to rank. In: DS 2019, 22nd International Conference on Discovery
  Science. pp. 279--289. No. 11828 in LNAI, Springer, Split, Croatia (2019)

\bibitem{ahmadi_huellermeier_aaai18}
{Ahmadi Fahandar}, M., H\"ullermeier, E.: Learning to rank based on analogical
  reasoning. In: Proceedings AAAI--2018, 32th AAAI Conference on Artificial
  Intelligence. pp. 2951--2958. New Orleans, Louisiana, USA (2018)

\bibitem{andr_sa95}
Andrews, R., Diederich, J., Tickle, A.: Survey and critique of techniques for
  extracting rules from trained artificial neural networks. Knowledge-Based
  Systems  \textbf{8}(6),  373--389 (1995)

\bibitem{beli_cb10}
Beliakov, G., James, S.: Citation-based journal ranks: the use of fuzzy
  measures. Fuzzy Sets and Systems  (2010)

\bibitem{app}
Bounhas, M., Pirlot, M., Prade, H.: Predicting preferences by means of
  analogical proportions. In: Case-Based Reasoning Research and Development.
  pp. 515--531. Springer International Publishing, Cham (2018)

\bibitem{BOUNHAS201736}
Bounhas, M., Prade, H., Richard, G.: Analogy-based classifiers for nominal or
  numerical data. International Journal of Approximate Reasoning  \textbf{91},
  36--55 (2017)

\bibitem{mpub173}
Cheng, W., H\"ullermeier, E.: Learning similarity functions from qualitative
  feedback. In: Bergmann, R., Althoff, K. (eds.) Proceedings ECCBR--2008, 9th
  European Conference on Case-Based Reasoning. pp. 120--134. Trier, Germany
  (2008)

\bibitem{cove_nn67}
Cover, T., Hart, P.: Nearest neighbor pattern classification. {\sc IEEE}
  Transactions on Information Theory  \textbf{IT-13},  21--27 (1967)

\bibitem{dubois16}
Dubois, D., Prade, H., Richard, G.: Multiple-valued extensions of analogical
  proportions. Fuzzy Sets and Systems  \textbf{292},  193--202 (2016)

\bibitem{mpub218}
F\"urnkranz, J., H\"ullermeier, E.: Preference Learning. Springer-Verlag (2011)

\bibitem{gent_tm89}
Gentner, D.: The mechanisms of analogical reasoning. In: Vosniadou, S., Ortony,
  A. (eds.) Similarity and Analogical Reasoning, pp. 197--241. Cambridge
  University Press (1989)

\bibitem{good_eu17}
Goodman, R., Flaxman, S.: European {U}nion regulations on algorithmic
  decision-making and a ``right to explanation''. AI Magazine  \textbf{38}(3),
  ~1--9 (2017)

\bibitem{guyo_ai03}
Guyon, I., Elisseeff, A.: An introduction to variable and feature selection.
  Journal of Machine Learning Research  \textbf{3},  1157--1182 (2003)

\bibitem{lund_au17}
Lundberg, S.M., Lee, S.I.: A unified approach to interpreting model
  predictions. In: Proc.\ NeurIPS, Advances in Neural Information Processing
  Systems. pp. 4765--4774 (2017)

\bibitem{Miclet2009}
Miclet, L., Prade, H.: Handling analogical proportions in classical logic and
  fuzzy logics settings. In: Proceedings ECSQARU, 10th European Conference on
  Symbolic and Quantitative Approaches to Reasoning with Uncertainty. pp.
  638--650. Springer Berlin Heidelberg, Berlin, Heidelberg (2009)

\bibitem{moln_im}
Molnar, C.: Interpretable Machine Learning: A Guide for Making Black Box Models
  Explainable (2018), \url{http://leanpub.com/interpretable-machine-learning}

\bibitem{ribe_ws16}
Ribeiro, M.T., Singh, S., Guestrin, C.: ``{W}hy should {I} trust you?''
  {E}xplaining the predictions of any classifier. In: Proceedings of the 22nd
  ACM SIGKDD International Conference on Knowledge Discovery and Data Mining.
  pp. 1135--1144 (2016)

\bibitem{same_ea}
Samek, W., Montavon, G., Vedaldi, A., Hansen, L., Müller, K.R. (eds.):
  Explainable AI: Interpreting, Explaining and Visualizing Deep Learning.
  Springer (2019)

\bibitem{vanl_ic19}
{Van Looveren}, A., Klaise, J.: Interpretable counterfactual explanations
  guided by prototypes. CoRR  \textbf{abs/1907.02584} (2019),
  \url{http://arxiv.org/abs/1907.02584}

\end{thebibliography}
	%

	\end{document}